\documentclass[11pt,twocolumn]{article}

\usepackage[top=0.9in, bottom=1in, left=0.35in, right=0.35in]{geometry}

\usepackage{graphicx}

\usepackage{amsmath}
\usepackage{amssymb}

\usepackage{multirow}
\usepackage{subcaption}

\usepackage{authblk}

\usepackage{url}

\begin{document}

	\title
	{
		Constructing Binary Descriptors with a Stochastic Hill Climbing Search
	}

	\author[$\dagger$]{Nenad Marku\v{s}}
	\author[$\dagger$]{Igor S. Pand\v{z}i\'{c}}
	\author[$\ddagger$]{J\"orgen Ahlberg}

	\affil[$\dagger$]
	{
		\normalsize
		University of Zagreb,
		Faculty of Electrical Engineering and Computing,
		Unska 3, 10000 Zagreb, Croatia
	}
	\affil[$\ddagger$]
	{
		Link\"{o}ping University,
		Department of Electrical Engineering,
		SE-581 83 Link\"{o}ping, Sweden
	}

	\date{}

	\maketitle

	\begin{abstract}
		Binary descriptors of image patches provide processing speed advantages and require less storage than methods that encode the patch appearance with a vector of real numbers.
		We provide evidence that, despite its simplicity, a stochastic hill climbing bit selection procedure for descriptor construction defeats recently proposed alternatives on a standard discriminative power benchmark.
		The method is easy to implement and understand, has no free parameters that need fine tuning, and runs fast.
	\end{abstract}

	\section{Introduction}
		Local image patch descriptors have become a widely used tool in computer vision, used for object/scene recognition \cite{visualwords}, image retrieval \cite{fisherkernel}, face verification \cite{fisherfaces}, face alignment \cite{sdm} and image stitching \cite{recognisingpanoramas}.
		Their usefulness and importance are proven by the large number of publications that introduced different descriptors.
		Recently binary keypoint descriptors \cite{brief,orb,brisk,dbrief,freak,binboost,ldb,rfd} gained considerable interest as they require less storage and provide faster matching times compared to descriptors that encode the patch appearance as a vector of real numbers \cite{sift,surf,lbp-desc,cvx}.

		In this paper, we investigate three bit selection procedures.
		Two of these have been recently used to construct discriminative local binary descriptors \cite{boostingtrick,binboost,ldb,rfd}: the boosting- and correlation-based methods.
		We show that a simple heuristic combinatorial search algorithm outperforms these approaches on a large discriminative power benchmark \cite{brown}.
		Additionally, we introduce a new descriptor based on binarised LBP features and compare it to competing approaches.

	\section{Selecting discriminative bits}\label{sec:sdb}
		A binary descriptor consists of $b$ classifiers.
		Each classifier, denoted as a \emph{bit} in further text, outputs a $0$ or a $1$ for a given image patch.
		Thus, a descriptor maps image patches into binary vectors which are used as signatures for search engines.
		The idea is to select individual bits in such a way that matching patches are "close" in the Hamming space and non-matching patches are "far".
		Let $\text{AUC}(d)$ denote the area under the receiver operating characteristics (ROC) curve for a descriptor $d$, measured on a set of matching and non-matching image patch pairs.
		The true positive rates (TPRs) and false positive rates (FPRs) are computed by thresholding the Hamming distance between signatures.
		Brown et al. \cite{brown} proposed to select the \emph{parameters} of real-valued descriptors (such as pooling region locations) by optimizing the AUC criterion.
		We apply this reasoning to select individual \emph{bits} (i.e., \emph{dimensions}) of a binary descriptor.
		Since the AUC criterion is not continuous, we apply stochastic hill climbing to achieve our goal.
		This also relates our paper to a large body of research in \emph{feature selection} (for example, see \cite{mrmr,fsel}).
		The whole procedure can be summarized by the following steps:
		\begin{enumerate}
			\item
				Generate a pool $P$ of $B$ bits.
			\item
				Select $b$ bits from $P$ to obtain a descriptor $d^*$.
			\item
				Iterate $N$ times:
				\begin{enumerate}
					\item
						Swap a random bit from $d^*$ with a random bit from $P$ to obtain $d$.
					\item
						If $\text{AUC}(d^*)<\text{AUC}(d)$, set $d^*=d$.
				\end{enumerate}
		\end{enumerate}
		The number of iterations, $N$, is set to $4\cdot B$ as this led to good results in all our experiments.
		Note that the described procedure does not specify the exact method how to generate individual bits to obtain a pool $P$.
		The method could be taken, for example, from \cite{rfd}.

		Note that optimizing the AUC criterion is not new in the machine learning community \cite{wmw}.
		Also, Lin et al. \cite{rankingmeasures} used the AUC criterion and other ranking measures to learn binary descriptors for image retrieval.
		The main difference between our approach and similar previous work \cite{brown,rankingmeasures} is that we eliminate the need for a complicated numerical solver by heavily exploiting randomization.
		We perform experimental validation of the proposed approach in the next sections.

	\section{Comparison of bit selection methods}
		We compare the proposed procedure to the recent boosting- and correlation-based bit selection methods.
		The boosting-based methods \cite{binboost,ldb} use the principle of reweighting the training data during learning, inspired by AdaBoost \cite{boosting}.
		The correlation-based method \cite{rfd} sequentially selects accurate bits that have low correlation with other already selected bits.
		Each method has a continuous parameter that significantly influences performance:
		the boosting shrinkage coefficient and the correlation threshold.
		In our experiments, we select these parameters to \emph{maximize} the accuracy on the test set.
		Note that the proposed descriptor construction procedure does not have free parameters that need to be fine-tuned. 

		We use the dataset introduced by Brown et al. \cite{brown} to provide experimental evidence that the proposed bit selection procedure has practical value.
		We report the results in terms of ROC curves and 95\% error rates.
		The dataset consists of three subsets: \emph{Notre Dame}, \emph{Liberty} and \emph{Yosemite}.
		Each contains a large number of $64\times 64$ rotation- and scale-normalized patches extracted around DoG keypoints \cite{sift}.
		The ground truth for each subset consists of $100$k, $200$k and $500$k pairs of patches, 50\% correspond to matching pairs and 50\% to non-matching pairs.
		We use a simplified notation for the training and testing subsets in our experiments.
		For example, L/ND will denote the scenario in which the \emph{Liberty} subset patch pairs were used for descriptor learning and the \emph{Notre Dame} subset patch pairs for testing.

\begin{figure*}
			\centering
			\begin{subfigure}[b]{0.45\textwidth}
				\includegraphics[width=\textwidth]{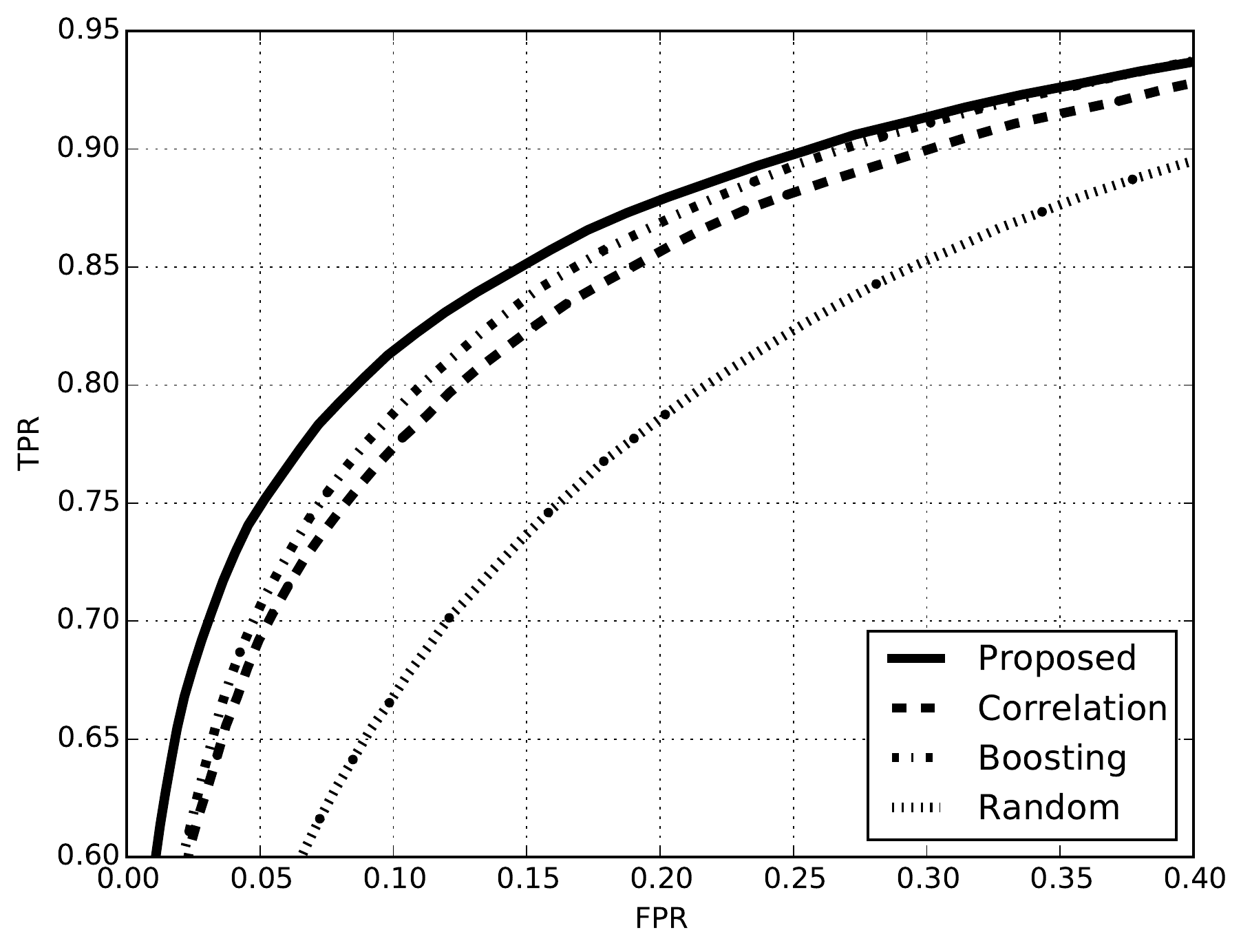}
				\caption{ND/L}
				\label{fig:nl}
			\end{subfigure}
			\begin{subfigure}[b]{0.45\textwidth}
				\includegraphics[width=\textwidth]{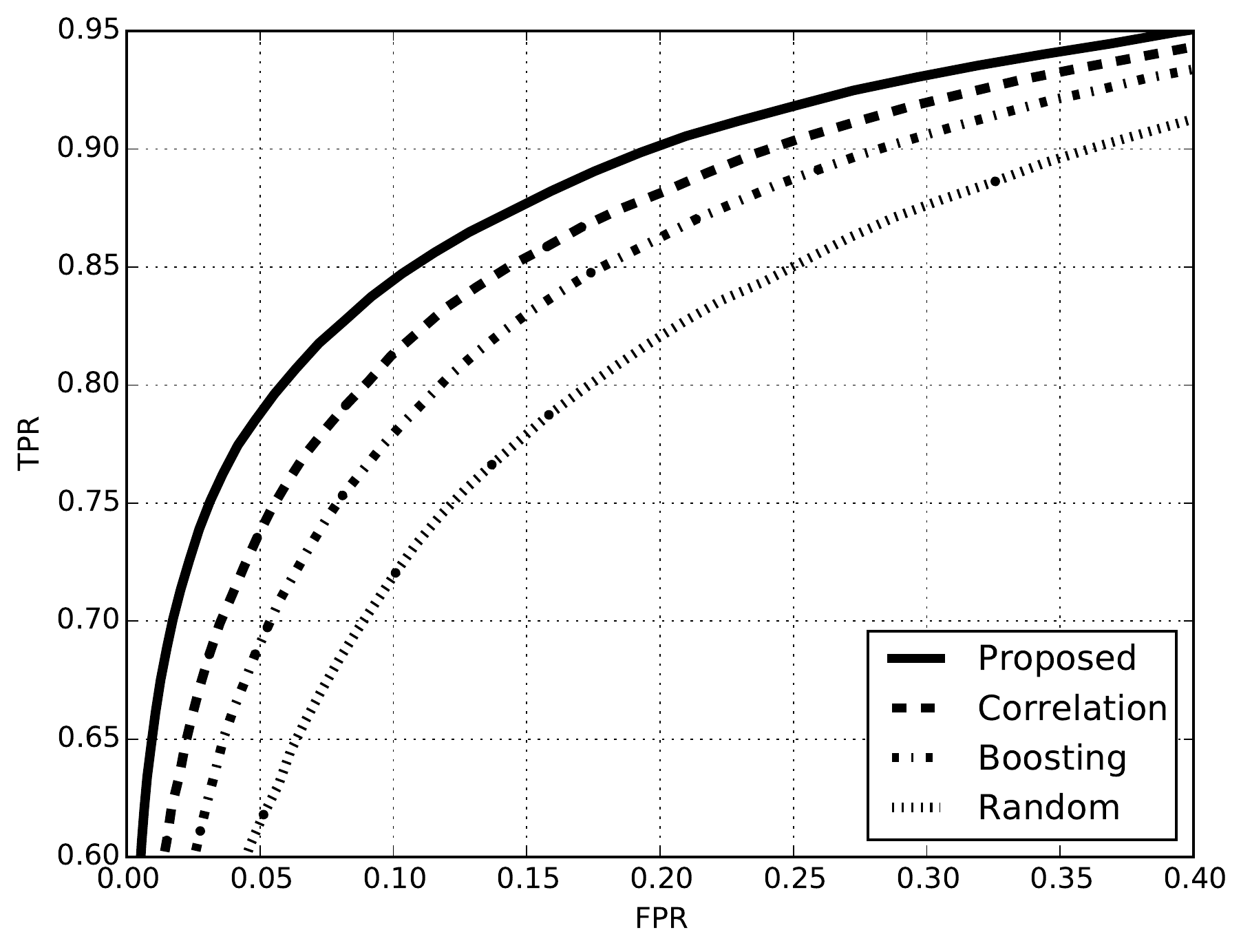}
				\caption{L/ND}
				\label{fig:ln}
			\end{subfigure}
			\begin{subfigure}[b]{0.45\textwidth}
				\includegraphics[width=\textwidth]{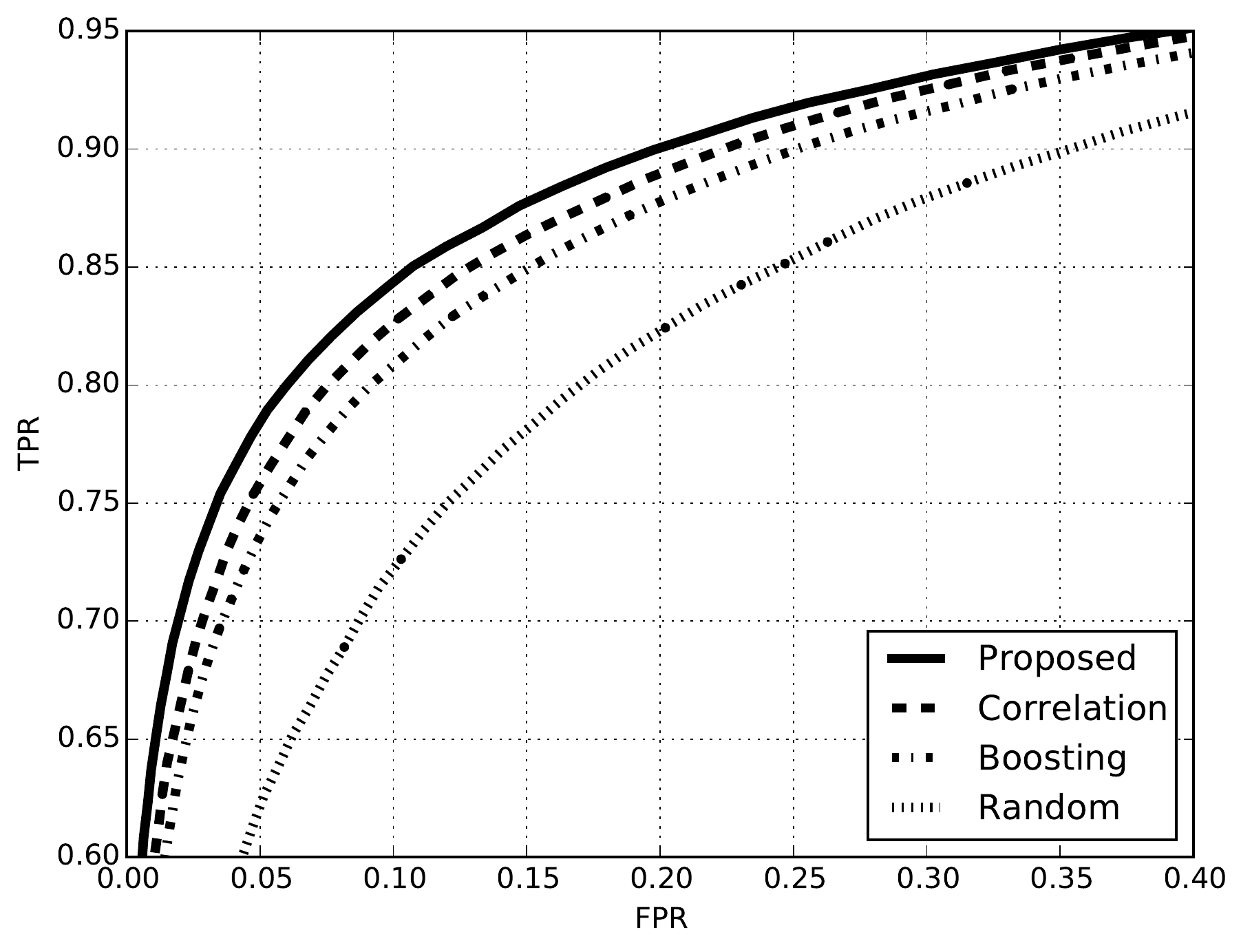}
				\caption{Y/ND}
				\label{fig:yn}
			\end{subfigure}
			\begin{subfigure}[b]{0.45\textwidth}
				\includegraphics[width=\textwidth]{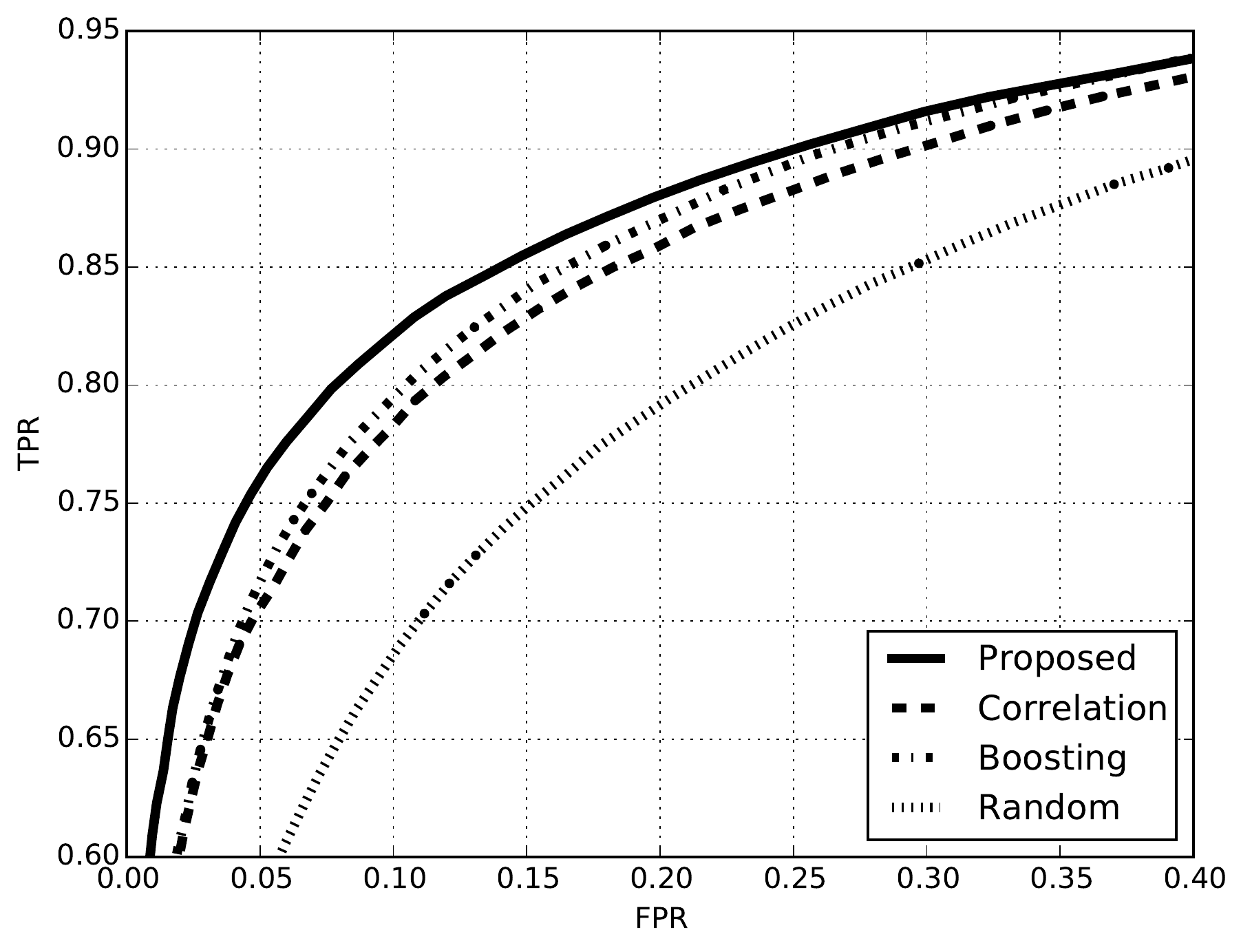}
				\caption{ND/Y}
				\label{fig:ny}
			\end{subfigure}
			\begin{subfigure}[b]{0.45\textwidth}
				\includegraphics[width=\textwidth]{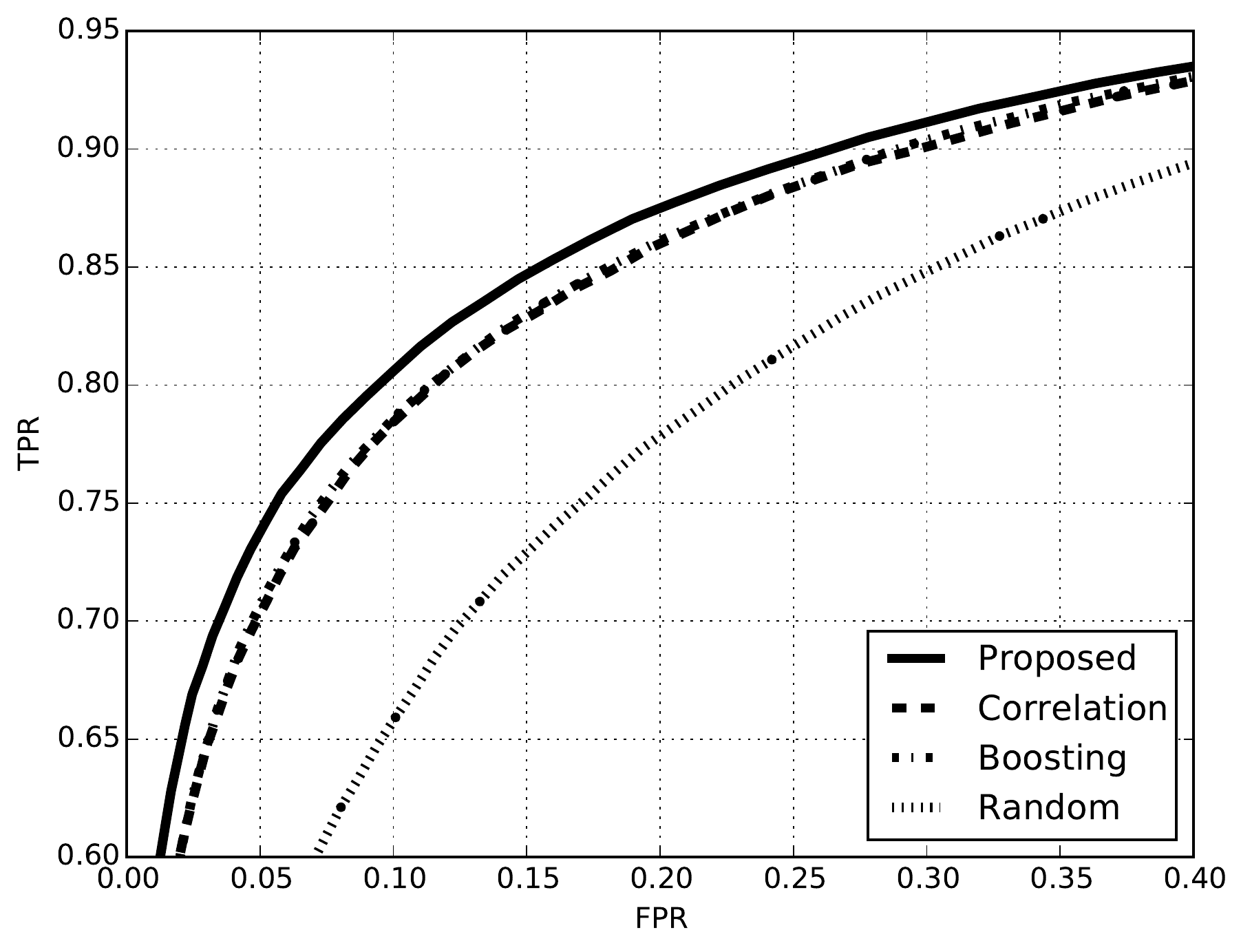}
				\caption{Y/L}
				\label{fig:yl}
				\end{subfigure}
			\begin{subfigure}[b]{0.45\textwidth}
				\includegraphics[width=\textwidth]{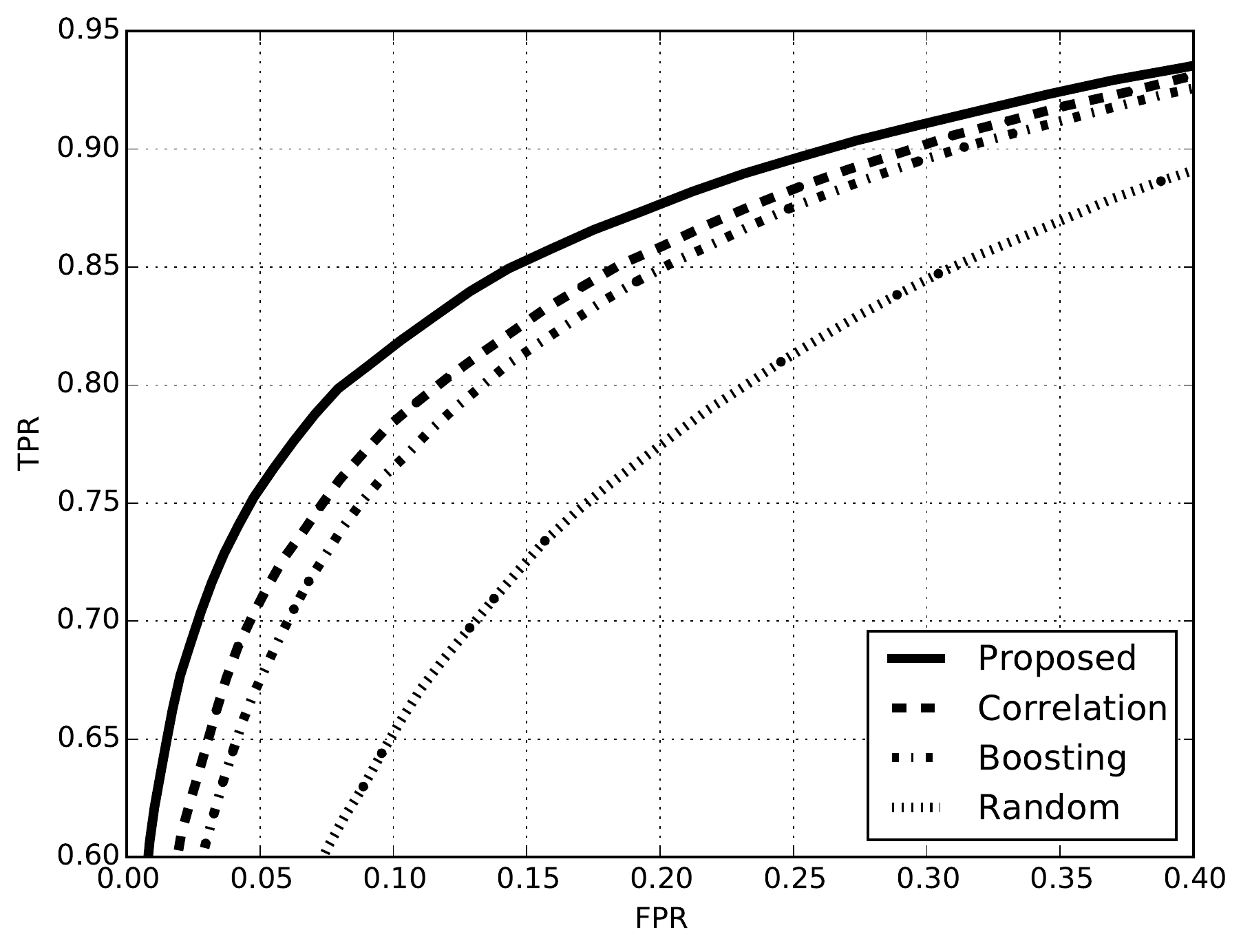}
				\caption{L/Y}
				\label{fig:ly}
			\end{subfigure}
			\caption
			{
				The ROC curves of the improved BRIEF descriptors on the testing subsets from \cite{brown}.
			}
			\label{fig:roc}
		\end{figure*}

		We compare the mentioned bit selection procedures on the task of improving the BRIEF descriptor \cite{brief}.
		The basic idea of BRIEF is to construct a $256$ bit descriptor by performing $256$ pixel intensity comparison binary tests ("$I_{x_1, y_1}<I_{x_2, y_2}?$") on an incoming image patch.
		The pixel sampling locations are fixed in runtime.
		It is intuitive that the distribution of these locations matters for accuracy.
		The authors of the original paper \cite{brief} propose to pick them at random.
		Here we show that the accuracy of the descriptor improves if we carefully select $256$ tests out of $1024$ random ones.
		This is done by comparing boosting-, correlation- or stochastic hill climbing-based bit selection procedures on training and test sets from \cite{brown}.
		Figure \ref{fig:roc} shows the resulting ROC curves, and Table \ref{tbl:acc} shows the 95\% error rates.
		
		\begin{table*}
			\center
			\resizebox{1.5\columnwidth}{!}
			{
				\begin{tabular}{|c||c||c|c|c|c|}
					\hline
					\multirow{2}{*}{Training subset} & \multirow{2}{*}{Testing subset} & \multicolumn{4}{|c|}{FPR at 95\% TPR} \\
					\cline{3-6}
					&	&	Random \cite{brief}	&	Boosting \cite{binboost,ldb}	&	Correlation \cite{rfd}	&	Proposed	\\
					\hline
					\hline
					\begin{tabular}{@{}c@{}}L\\Y\end{tabular}	&	ND	&	$55.71\pm 0.91$	&	\begin{tabular}{@{}c@{}}$47.89$\\$44.69$\end{tabular}	&	\begin{tabular}{@{}c@{}}$43.62$\\$41.21$\end{tabular}	&	\begin{tabular}{@{}c@{}}$\mathbf{39.26\pm 0.29}$\\$\mathbf{39.24\pm 0.32}$\end{tabular}	\\
					\hline
					\begin{tabular}{@{}c@{}}ND\\Y\end{tabular}	&	L	&	$61.34\pm 0.70$	&	\begin{tabular}{@{}c@{}}$\mathbf{46.49}$\\$49.92$\end{tabular}	&	\begin{tabular}{@{}c@{}}$51.11$\\$51.21$\end{tabular}	&	\begin{tabular}{@{}c@{}}$47.05\pm 0.26$\\$\mathbf{47.89\pm 0.42}$\end{tabular}	\\
					\hline
					\begin{tabular}{@{}c@{}}ND\\L\end{tabular}	&	Y	&	$60.94\pm 1.11$	&	\begin{tabular}{@{}c@{}}$\mathbf{45.51}$\\$51.58$\end{tabular}	&	\begin{tabular}{@{}c@{}}$48.92$\\$49.43$\end{tabular}	&	\begin{tabular}{@{}c@{}}$45.85\pm 0.25$\\$\mathbf{47.46\pm 0.24}$\end{tabular}	\\
					\hline
				\end{tabular}
			}
			\caption
			{
				Accuracy results for the improved BRIEF descriptor on subsets from \cite{brown}.
				The mean and standard deviation for methods that involve randomization were computed from $10$ runs.
			}
			\label{tbl:acc}
		\end{table*}
		We can see that the proposed method leads to lower error rates in the majority of scenarios.
		Also, its behavior is consistent, i.e., it obtains good results for all train/test pairs, which is not the case for the competing approaches.
		Bit selection processing times can be seen in Table~\ref{tbl:times}.
		\begin{table}
			\center
			\resizebox{\columnwidth}{!}
			{
				\begin{tabular}{| l || c | c | c |}
					\hline
						Selection type	&	Correlation \cite{rfd}	&	Boosting \cite{binboost,ldb}	&	Proposed	\\
					\hline
						Time [s]	&	$3281$	&	$927$	&	$175$	\\
					\hline
				\end{tabular}
			}
			\caption
			{
				Processing times for selecting $256$ out of $1024$ bits for a dataset of $100$k patch pairs.
			}
			\label{tbl:times}
		\end{table}
		The parameters of all methods were fixed.
		Thus, this is not an entirely fair comparison since both boosting- and correlation-based approaches typically require several reruns for parameter fine-tuning.

		The proposed procedure also achieves better results when individual bits are based on binarised gradients \cite{binboost,rfd}.
		We omit these results for brevity.

		In the next section we introduce our new keypoint descriptor (generated with the proposed stochastic hill climbing bit selection procedure) and compare it to the state-of-the-art on the problem of visual object search.

		\section{Binarising LBP histograms}
			Local binary patterns (LBPs) are a well known computer vision tool.
			Heikkil\"a et al. \cite{lbp-desc} showed that their histograms can be used for keypoint description
			(the authors report results that were superior to SIFT).
			First, we use their procedure to extract a real-valued vector $\mathbf{v}=(v_1, v_2, \ldots, v_n)\in\mathbb{R}^n$ from each image patch.
			Next, we generate a large number of bits of the following form:
			$$
				\text{bit}_{i, j}(\mathbf{v})
				=
				\begin{cases}
					1,	&	v_i > v_j	\\
					0,	&	\text{otherwise}
				\end{cases}
			$$
			(the tuple of two indices, $(i, j)$, specifies the bit; $v_i$ and $v_j$ are the $i$-th and $j$-th components of $\mathbf{v}$).
			Finally, we run the proposed stochastic hill climbing procedure on the dataset of Brown et al. \cite{brown} to select $256$ discriminative bits which will form our binarised LBP descriptor (bLBP).

			In the next subsection we compare the proposed descriptor with recently introduced competition.

			\subsection{Experiments in image retrieval}
			Following \cite{rfd}, we implemented a simple visual search system to compare the discriminative power of different descriptors.
			Each image is represented with $75$ FAST \cite{fast} keypoints.
			The  retrieval  is  based  on  the number of matching keypoints between the query and database images.
			Whether two keypoints match or not is determined by thresholding the Hamming distance of their descriptors.
			The threshold is fine-tuned for each descriptor type to give best retrieval results on a given database.
			We use the following databases:
			\begin{itemize}
				\item
					UKB \cite{ukb} (first $300$ objects, $4$ views each)
				\item
					ZuBuD \cite{zubud} ($200$ buildings, $5$ images each)
				\item
					COIL-100 \cite{coil-100} ($100$ objects, first $32$ views each)
				\item
					INRIA Holidays \cite{holidays}
					(approximately $1500$ images of $500$ different scenes)
			\end{itemize}
			For each image in the current database we search for the $k$ most similar ones among the remaining images.
			We report the average precision at these top $k$ results as retrieval accuracy.	
			The integer $k$ is specific for each database.
			For example, we use $k=3$ for UKB since it contains $4$ views per object and $k=1$ for INRIA Holidays since there is a variable number of images for each scene.

			Table \ref{tbl:comparison} summarizes the obtained retrieval results
			(we used the descriptor extraction code provided by the authors to have a fair comparison).
			\begin{table*}
				\center
				\resizebox{2\columnwidth}{!}
				{
					\begin{tabular}{|c|c||c|c|c|c|}
						\hline
							\multirow{2}{*}{Descriptor} & \multirow{2}{*}{Size} & \multicolumn{4}{|c|}{Average precision@$k$ [\%]} \\
							\cline{3-6}
							&	&	UKB (testing) \cite{ukb}, $k=3$	&	ZuBuD \cite{zubud}, $k=4$	&	COIL-100 \cite{coil-100}, $k=31$	&	INRIA Holidays \cite{holidays}, $k=1$	\\
						\hline
						\hline
						bLBP	&	$256$b	&	$58.9\pm 1.9$	&	$68.5\pm 0.8$	&	$48.0\pm 0.8$	&	$42.6\pm 1.3$	\\
						\hline
						RFD-G \cite{rfd}	&	$448$b	&	$58.3$	&	$73.7$	&	$44.0$	&	$45.9$	\\
						\hline
						RFD-R \cite{rfd}	&	$320$b	&	$51.0$	&	$68.4$	&	$43.6$	&	$44.1$	\\	
						\hline
						BinBoost256 \cite{binboost}	&	$256$b	&	$43.6$	&	$67.3$	&	$37.9$	&	$36.5$	\\
						\hline
						LDB \cite{ldb}	&	$256$b	&	$41.2$	&	$54.9$	&	$33.2$	&	$42.2$	\\
						\hline
						BRIEF \cite{brief}	&	$256$b	&	$38.7$	&	$59.3$	&	$24.4$	&	$31.9$	\\
						\hline
					\end{tabular}
				}
				\caption
				{
					Comparison of the proposed approach (bLBP) to other relevant methods from the literature.
				}
				\label{tbl:comparison}
			\end{table*}
			We can see that the proposed descriptor outperforms all competing approaches with size of $256$ bits.
			Also, its results are comparable or better to the ones obtained by RFD \cite{rfd}, which requires more storage ($320$ or $448$ bits).
			Table \ref{tbl:speed} shows the processing speed obtained by each of the methods.
			\begin{table}
				\center
				\resizebox{0.6\columnwidth}{!}
				{
					\begin{tabular}{| c || c |}
						\hline
							Descriptor	&	Time [ms]	\\
						\hline
							bLBP	&	$5.1$	\\
						\hline
							RFD-G \cite{rfd}	&	$243$	\\
						\hline
							RFD-R \cite{rfd}	&	$30.5$	\\
						\hline
							BinBoost256 \cite{binboost}	&	$27.6$	\\
						\hline
							LDB \cite{ldb}	&	$0.43$	\\
						\hline
							BRIEF \cite{brief}	&	$0.1$	\\
						\hline
					\end{tabular}
				}
				\caption
				{
					Time in miliseconds needed to extract descriptors from $75$ keypoints for different approaches.
				}
				\label{tbl:speed}
			\end{table}

	\section{Conclusion}
		We have shown that a simple stochastic hill climbing bit selection procedure outperforms recent alternatives \cite{binboost,ldb,rfd} on a standard dataset \cite{brown}.
		We also introduced a new binary desciptor based on binarised LBP features that achieved good results in terms of accuracy and processing speed when compared to competing approaches.
		The source code is available at \url{http://public.tel.fer.hr/bitslkt/}.

	{
		\section*{Acknowledgements}
			This research is partially supported by Visage Technologies AB, Link\"oping, Sweden, and by the Ministry of Science, Education and Sports of the Republic of Croatia. 
	}

	\small
	{
		\bibliographystyle{plain}
		\bibliography{references}

\begin{thebibliography}{10}

\bibitem{freak}
A.~Alahi, R.~Ortiz, and P.~Vandergheynst.
\newblock {{FREAK}: Fast Retina Keypoint}.
\newblock In {\em CVPR}, 2012.

\bibitem{surf}
H.~Bay, T.~Tuytelaars, and L.~Van Gool.
\newblock {{SURF}: Speeded Up Robust Features}.
\newblock In {\em ECCV}, 2006.

\bibitem{brown}
M.~Brown, G.~Hua, and S.~Winder.
\newblock {Discriminative Learning of Local Image Descriptors}.
\newblock {\em PAMI}, 2011.

\bibitem{recognisingpanoramas}
M.~Brown and D.~G. Lowe.
\newblock {Recognising panoramas}.
\newblock In {\em ICCV}, 2003.

\bibitem{brief}
M.~Calonder, V.~Lepetit, C.~Strecha, and P.~Fua.
\newblock {{BRIEF}: Binary Robust Independent Elementary Features}.
\newblock In {\em ECCV}, 2010.

\bibitem{visualwords}
G.~Csurka, C.~Dance, L.~Fan, J.~Willamowski, and C.~Bray.
\newblock {Visual categorization with bags of keypoints}.
\newblock In {\em ECCV}, 2004.

\bibitem{rfd}
B.~Fan, Q.~Kong, T.~Trzcinski, Z.~Wang, C.~Pan, and P.~Fua.
\newblock {Receptive Fields Selection for Binary Feature Description}.
\newblock {\em IEEE Transaction on Image Processing}, 2014.

\bibitem{boosting}
J.~Friedman, T.~Hastie, and R.~Tibshirani.
\newblock {Additive Logistic Regression: a Statistical View of Boosting}.
\newblock {\em Annals of Statistics}, 1998.

\bibitem{fsel}
I.~Guyon and A.~Elisseeff.
\newblock {An Introduction to Variable and Feature Selection}.
\newblock {\em JMLR}, 2003.

\bibitem{lbp-desc}
M.~Heikkil\"a, M.~Pietik\"ainen, and C.~Schmid.
\newblock Description of interest regions with local binary patterns.
\newblock {\em Pattern Recognition}, 2009.

\bibitem{holidays}
H.~Jegou, M.~Douze, and C.~Schmid.
\newblock Hamming embedding and weak geometric consistency for large scale
  image search.
\newblock In {\em ECCV}, 2008.

\bibitem{brisk}
S.~Leutenegger, M.~Chli, and R.~Y. Siegwart.
\newblock {{BRISK}: Binary Robust Invariant Scalable Keypoints}.
\newblock In {\em ICCV}, 2011.

\bibitem{rankingmeasures}
G.~Lin, C.~Shen, and J.~Wu.
\newblock {Optimizing Ranking Measures for Compact Binary Code Learning}.
\newblock In {\em ECCV}, 2014.

\bibitem{sift}
D.~G. Lowe.
\newblock {Object recognition from local scale-invariant features}.
\newblock In {\em ICCV}, 1999.

\bibitem{coil-100}
S.~A. Nene, S.~K. Nayar, and H.~Murase.
\newblock Columbia object image library (coil-100).
\newblock Technical report, Columbia University, 1996.

\bibitem{ukb}
D.~Nist\'er and H.~Stew\'enius.
\newblock Scalable recognition with a vocabulary tree.
\newblock In {\em CVPR}, 2006.

\bibitem{mrmr}
H.~Peng, F.~Long, and C.~Ding.
\newblock {Feature Selection Based on Mutual Information: Criteria of
  Max-Dependency, Max-Relevance, and Min-Redundancy}.
\newblock {\em PAMI}, 2005.

\bibitem{fisherkernel}
F.~Perronnin, J.~Sanchez, and T.~Mensink.
\newblock {Improving the Fisher Kernel for Large-Scale Image Classification}.
\newblock In {\em ECCV}, 2010.

\bibitem{fast}
E.~Rosten, R.~Porter, and T.~Drummond.
\newblock Faster and better: A machine learning approach to corner detection.
\newblock {\em PAMI}, 2008.

\bibitem{orb}
E.~Rublee, V.~Rabaud, K.~Konolige, and G.~R. Bradski.
\newblock {{ORB}: An efficient alternative to SIFT or SURF}.
\newblock In {\em ICCV}, 2011.

\bibitem{zubud}
H.~Shao, T.~Svoboda, and L.~V. Gool.
\newblock Zubud---z\"urich building database for image based recognition.
\newblock Technical report, ETH Z\"urich, 2003.

\bibitem{fisherfaces}
K.~Simonyan, O.~M. Parkhi, A.~Vedaldi, and A.~Zisserman.
\newblock {Fisher Vector Faces in the Wild}.
\newblock In {\em BMVC}, 2013.

\bibitem{cvx}
K.~Simonyan, A.~Vedaldi, and A.~Zisserman.
\newblock {Learning Local Feature Descriptors Using Convex Optimisation}.
\newblock In {\em ECCV}, 2012.

\bibitem{boostingtrick}
T.~Trzcinski, M.~Christoudias, V.~Lepetit, and P.~Fua.
\newblock {Learning Image Descriptors with the Boosting-Trick}.
\newblock In {\em NIPS}, 2012.

\bibitem{binboost}
T.~Trzcinski, M.~Christoudias, V.~Lepetit, and P.~Fua.
\newblock {Boosting Binary Keypoint Descriptors}.
\newblock In {\em CVPR}, 2013.

\bibitem{dbrief}
T.~Trzcinski and V.~Lepetit.
\newblock {Efficient Discriminative Projections for Compact Binary
  Descriptors}.
\newblock In {\em ECCV}, 2012.

\bibitem{sdm}
X.~Xiong and F.~De la~Torre.
\newblock {Supervised Descent Method and its Applications to Face Alignment}.
\newblock In {\em CVPR}, 2013.

\bibitem{wmw}
L.~Yan, R.~Dodier, M.~C. Mozer, and R.~Wolniewicz.
\newblock Optimizing classifier performance via an approximation to the
  wilcoxon-mannwhitney statistic.
\newblock In {\em ICML}, 2003.

\bibitem{ldb}
X.~Yang and K.-T. Cheng.
\newblock {Local Difference Binary for Ultra-fast and Distinctive Feature
  Description}.
\newblock {\em PAMI}, 2014.

\end{thebibliography}
	}

\end{document}